\documentclass[10pt,twocolumn,letterpaper]{article}

\usepackage{cvpr}
\usepackage{times}
\usepackage{epsfig}
\usepackage{graphicx}
\usepackage{amsmath}
\usepackage{amssymb}
\usepackage{multirow}
\usepackage{multicol}
\usepackage{hhline}
\usepackage{tabularx}
\usepackage{booktabs}
\usepackage[breaklinks=true,bookmarks=false]{hyperref}

\cvprfinalcopy 
\def\cvprPaperID{5458} 

\ifcvprfinal\fi

\newcolumntype{C}{>{\centering\arraybackslash}p{3em}}
\newcolumntype{R}{>{\raggedleft\arraybackslash}p{3em}}

\begin{document}

\title{Attribute-aware Face Aging with Wavelet-based Generative Adversarial Networks}

\author{Yunfan Liu$^{1}$\thanks{Authors contributed equally.}\qquad Qi Li$^{1,2,3}$\footnotemark[1]\qquad Zhenan Sun$^{1,2,4}$\\
$^{1}$ Center for Research on Intelligent Perception and Computing, CASIA\\
$^{2}$ National Laboratory of Pattern Recognition, CASIA\\
$^{3}$ Artificial Intelligence Research, CAS, Jiaozhou, Qingdao, China\\
$^{4}$ Center for Excellence in Brain Science and Intelligence Technology, CAS\\
{\tt\small yunfan.liu@cripac.ia.ac.cn, \{qli, znsun\}@nlpr.ia.ac.cn}
}

\maketitle
\thispagestyle{empty}

\begin{abstract}
   Since it is difficult to collect face images of the same subject over a long range of age span, most existing face aging methods resort to unpaired datasets to learn age mappings.
   However, the matching ambiguity between young and aged face images inherent to unpaired training data may lead to unnatural changes of facial attributes during the aging process, which could not be solved by only enforcing identity consistency like most existing studies do.
   In this paper, we propose an attribute-aware face aging model with wavelet-based Generative Adversarial Networks (GANs) to address the above issues.
   To be specific, we embed facial attribute vectors into both the generator and discriminator of the model to encourage each synthesized elderly face image to be faithful to the attribute of its corresponding input.
   In addition, a wavelet packet transform (WPT) module is incorporated to improve the visual fidelity of generated images by capturing age-related texture details at multiple scales in the frequency space. 
   Qualitative results demonstrate the ability of our model in synthesizing visually plausible face images, and extensive quantitative evaluation results show that the proposed method achieves state-of-the-art performance on existing datasets.
\end{abstract}

\section{Introduction}

Face aging, also known as age progression~\cite{shu2015personalized}, aims at rendering a given face image with aging effects while still preserving personalized features. 
Applications of face aging techniques range from social security to digital entertainment, such as predicting contemporary appearance of missing children and cross-age identity verification. 
Due to the practical value of face aging, many approaches have been proposed to address this problem in the last two decades~\cite{lanitis2002toward,tazoe2012facial,suo2010compositional,tiddeman2001prototyping,kemelmacher2014illumination}.
With the rapid development of deep learning, deep generative models are widely adopted to synthesize aged face images~\cite{wang2016recurrent,duong2016longitudinal,duong2017temporal}. 
However, the most critical problem of these methods is that multiple face images of the same person at different ages are required at training stage, which is extremely expensive to collect in practice and thus their applications are largely limited.

\begin{figure}[t]
\begin{center}
\includegraphics[width=1.0\linewidth]{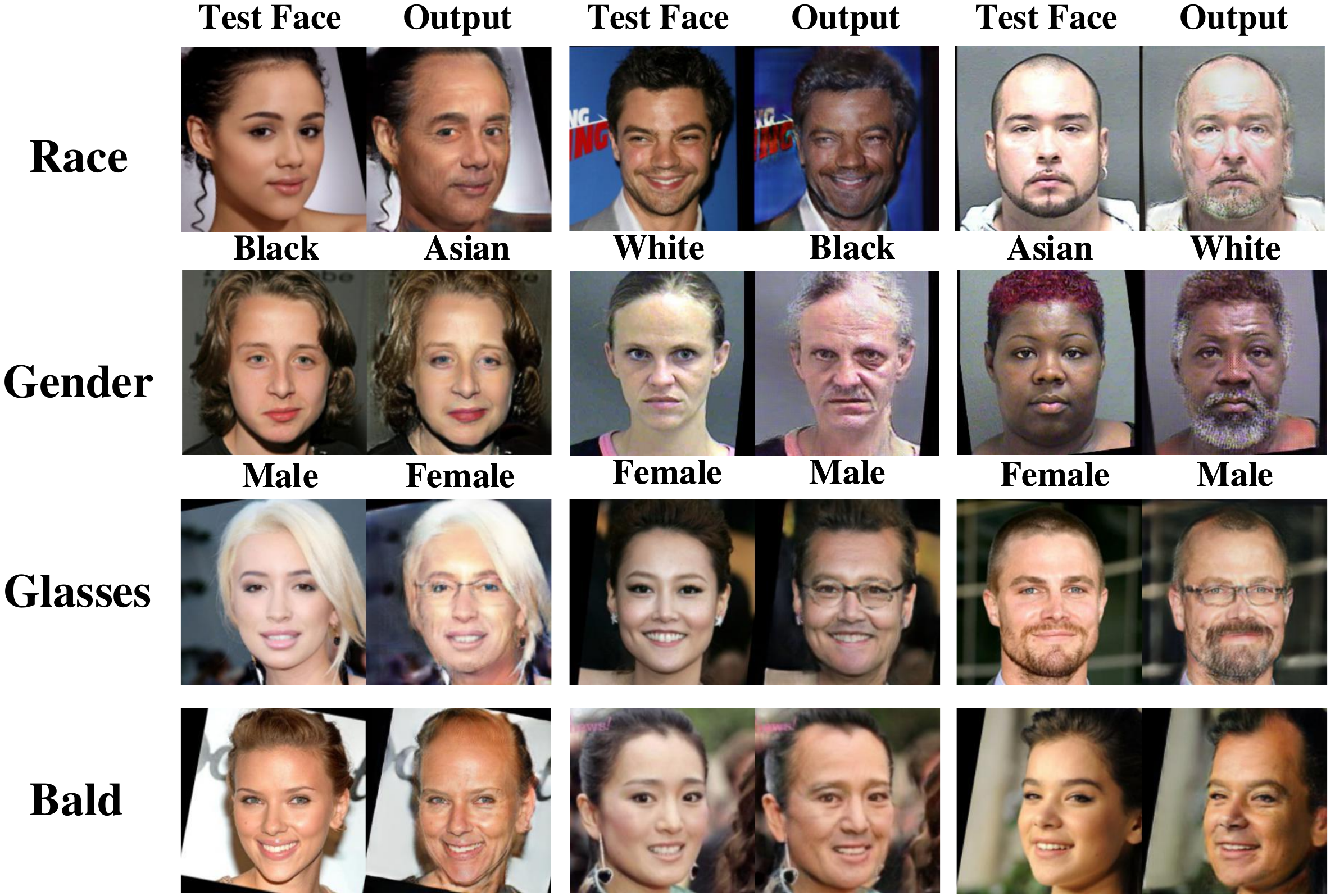}
\end{center}
\caption{Examples of face aging with mismatched facial attributes generated by face aging model without facial attribute embedding. Four attributes (Race, Gender, Glasses, and Bald) are considered and three sample results are presented for each. Labels of `Race' and `Gender' are all obtained via advanced publicly available APIs of Face++~\cite{face2018toolkit} and placed underneath each image.}
\label{fig:faceAttIdt}
\end{figure}

To deal with this problem, many recent studies resort to unpaired face aging data to train the model~\cite{wang2016recurrent,zhang2017age,yang2017learning,li2018global}.
However, these approaches mainly focus on face aging itself while neglecting other critical conditional information of the input (\eg, facial attributes), thus fail to regulate the training process. 
Consequently, training face image pairs with mismatched attributes would mislead the model into learning translations other than aging, causing serious ghosting artifacts and even incorrect facial attributes in generation results.
Fig.~\ref{fig:faceAttIdt} shows some face aging results with mismatched attributes.
In the rightmost face aging result under `gender', beard is mistakenly attached to the input female face image. 
This is because the model learns that growing a beard is a typical sign of aging but has no way to know that this does not happen to a woman, as face image pairs of young woman and old man could be treated as positive training samples if no conditional information is considered.

To suppress such undesired changes of semantic information during the aging process, many recent face aging studies attempt to supervise the output by enforcing identity consistency~\cite{zhang2017age,antipov2017face,yang2017learning,li2018global}. 
However, as shown in Fig.~\ref{fig:faceAttIdt}, personalized features are well preserved in the output for all sample results, nevertheless, obvious unnatural changes of facial attributes are still observed.
In other words, well maintained identity-related features do NOT imply reasonable aging results when training with unpaired data. Therefore, merely enforcing identity consistency is insufficient to eliminate matching ambiguities in unpaired training data, thus fails to achieve satisfactory face aging performance.

To solve the above-mentioned issues, in this paper, we propose a framework based on generative adversarial networks (GANs). 
Different from existing methods in the literature, we involve semantic conditional information of the input by embedding facial attribute vectors in both the generator and discriminator, so that the model could be guided to output elderly face images with attributes faithful to each corresponding input.
Furthermore, to enhance aging details, based on the observation that signs of aging are mainly represented by wrinkles, laugh lines, and eye bags, which could be treated as local textures, we employ wavelet packet transform to extract features at multiple scales in the frequency space efficiently. 

To summarized, the main contributions are as follows:
\begin{itemize}
\item Facial attributes are incorporated as conditional information into both the generator and discriminator for face aging, since identity preservation is insufficient for generating reasonable results. 

\item Wavelet packet transform is adopted to extract features of texture details at multiple scales in the frequency domain for generating fine-grained details of aging effects.

\item Extensive experiments have been conducted to demonstrate the ability of the proposed method in rendering accurate aging effects and preserving information of both identity and facial attributes. Quantitative results indicate that our method achieves state-of-the-art performance.
\end{itemize}

\section{Related Work}
In the last few decades, face aging has been a very popular research topic and a great amount of algorithms have been proposed to tackle this issue. 
In general, these methods could be divided into three categories: physical model-based methods, prototype-based methods, and deep learning-based methods.

Physical model-based methods mechanically simulate the changes of facial appearance~\wrt time by modeling the anatomical structure of human faces. 
Todd et al.~\cite{todd1980perception} modeled the translation of facial appearance by revised cardioidal strain transformation. 
Subsequent works investigated the problem from various biological aspects including muscles and overall facial structures~\cite{lanitis2002toward,tazoe2012facial}. However, physical model-based algorithms are computational expensive and large amount of image sequences of the same subject are required to model aging effects.

Data-driven prototyping approaches~\cite{suo2010compositional,tiddeman2001prototyping,kemelmacher2014illumination} come into view the next, where faces are divided into age groups and each group is represented by an average face (prototype) computed from the training data. After that, translation patterns between prototypes are regarded as effects of aging. 
The main problem of prototyping methods is that personalized features are eliminated when calculating average faces, thus the identity information is not well preserved. 

In recent years, deep generative models with temporal architectures are adopted to synthesize images of elderly faces~\cite{wang2016recurrent,duong2016longitudinal,duong2017temporal}. However, in most of these works, face image sequence over a long age span for each subject is required thus their potential in practical use is limited. 
With the success of GANs~\cite{goodfellow2014generative} in generating visually appealing images, many efforts have been made to tackle the problem of face aging using GAN-based framework~\cite{zhang2017age,yang2017learning,li2018global,Song2018Dual,wang2018face_aging,Qi2018attention}. 
Zhang~\etal~\cite{zhang2017age} proposed a conditional adversarial autoencoder (CAAE) to achieve age progression and regression by traversing in low-dimensional manifold. 
The work most similar to ours is~\cite{yang2017learning}, in which a GAN-based model with pyramid architecture is proposed, and identity loss is adopted to achieve permanence. 
Besides preserving identity information, we focus on alleviating the influence of matching ambiguity of unpaired training samples and ensuring attribute consistency by embedding facial attribute vectors in the model.

\begin{figure*}[ht]
\begin{center}
\includegraphics[width=0.75\linewidth]{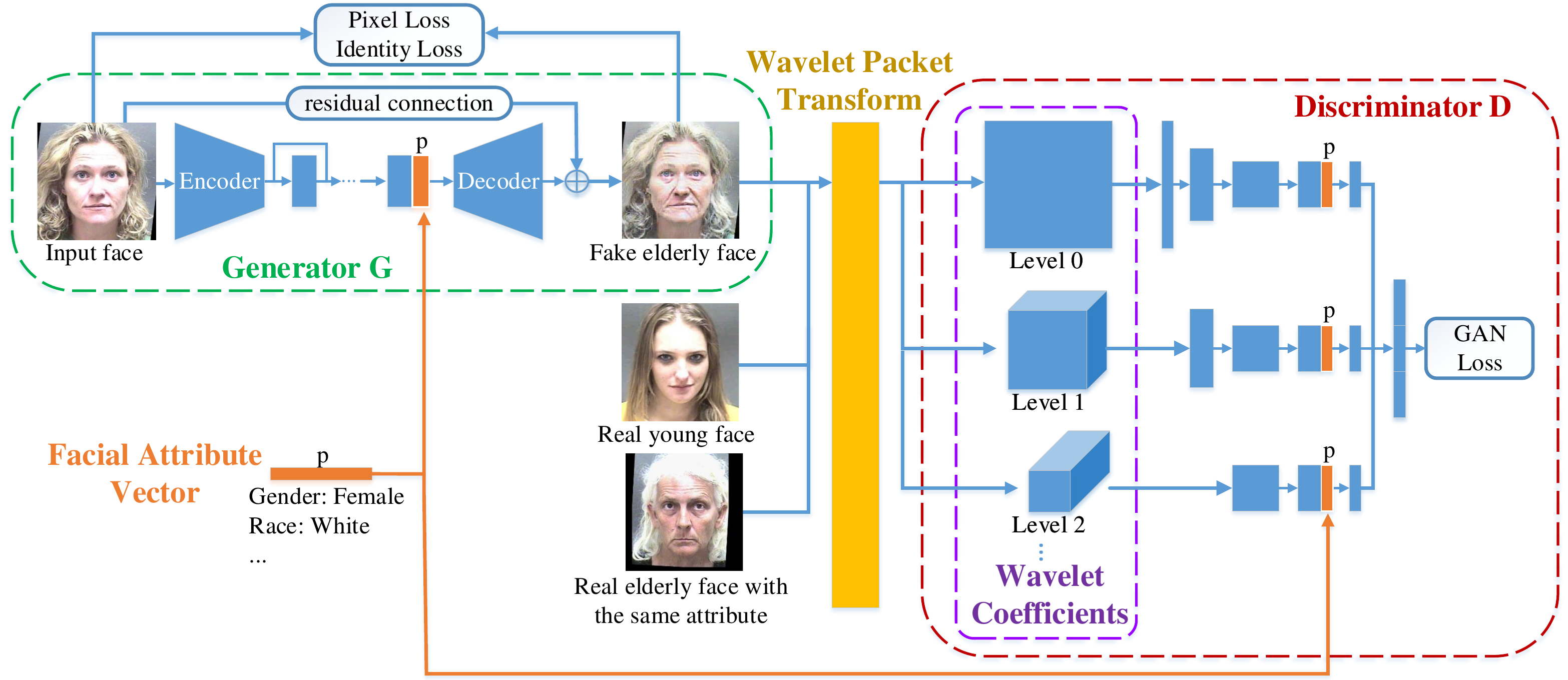}
\end{center}
\caption{An overview of the proposed face aging framework. An hourglass-shaped generator $G$ learns the age mapping and outputs lifelike elderly face images. A discriminator $D$ is employed to distinguish synthesized face images from generic ones, based on multi-scale wavelet coefficients computed by the wavelet packet transform module. The $p$-dimensional attribute vector describing the input face image is embedded to both the generator and discriminator to reduce matching ambiguity inherent to unpaired training data.}
\label{fig:overview}
\end{figure*}

\section{Approach}
In a unpaired face aging dataset, each young face image might map to many elderly face candidates during the training process, and image pairs with mismatched semantic information may mislead the model into learning translations other than aging.
To solve this problem, we present a GAN-based face aging model that takes young face images and their semantic information (\ie facial attributes) as input and outputs visually plausible aged faces accordingly.
The network consists of two parts: a facial attribute embedded generator $G$ and a wavelet-based discriminator $D$. The generator network embeds facial attributes into young face images and synthesizes aged faces. 
The discriminator network is used to encourage the generation results to be indistinguishable from generic ones and to possess attributes same as the corresponding input.
An overview of the proposed framework is presented in Fig.~\ref{fig:overview}.

\subsection{Facial Attribute Embedded Generator}
Existing face aging studies~\cite{li2018global,yang2017learning,zhang2017age} only take young face images as inputs and then directly learn age mappings using GAN-based networks.
Although constraints on identity information and pixel values are usually imposed to restrict modifications made to input images, facial attributes may still undergo unnatural translations (as shown in Fig.~\ref{fig:faceAttIdt}).
Unlike previous works, we propose to incorporate both low-level image information (pixel values) and high-level semantic information (facial attributes) into the face aging model to regularize image translation patterns and reduce the ambiguity of mappings between unpaired young and aged faces.
To be specific, the model takes young face images and their corresponding attribute vectors as input, and generates elderly face images with attributes in agreement with the input ones. 

Rather than supervising the attributes of generation results by simply adopting an additional loss term, we embed the attribute vector in the generator so that semantic facial information is well considered in the generation process and encourages the model to produce face images with consistent attributes more effectively.
To be specific, we employ an hourglass-shaped fully convolutional network as the generator, which has achieved success in previous image translation studies~\cite{johnson2016perceptual,zhu2017unpaired}. It consists of an encoder network, a decoder network, and four residual blocks in between as the bottleneck.
The input facial attribute vector is replicated and concatenated to the output blob of the last residual block as they both contain high-level semantic features. After the combination, the decoder network transforms the concatenated feature blob back to the image space. 

Since face aging could be considered as rendering aging effects conditioned on the input young face image, we add the input image to the output of the decoder to form a residual connection. 
Compared to synthesize the whole face image, this structure automatically makes the generator focus more on modeling the difference between input and output face images, namely the representative signs of aging, and be less likely to be distracted by visual content irrelevant to aging, such as background.
Finally, the numeric scale of the resultant tensor is normalized by a hyperbolic tangent (tanh) mapping and thus the generated elderly face image is obtained.

\begin{figure}[t]
\centering\includegraphics[width=0.70\linewidth]{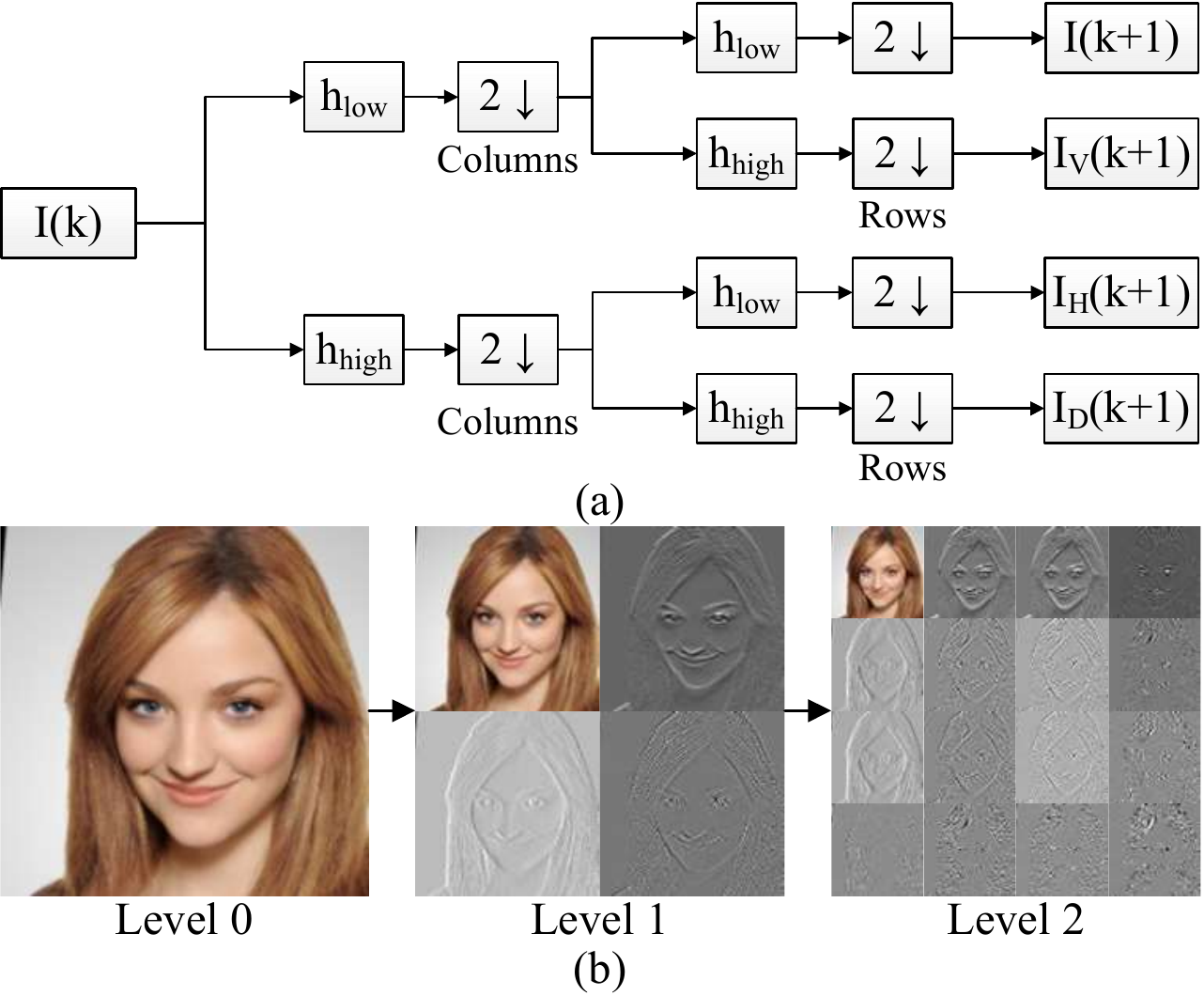}
\caption{Demonstration of wavelet packet transform. (a) Low-pass and high-pass decomposition filters ($h_{low}$ and $h_{high}$) are applied iteratively to the input on $k$-th level to compute wavelet coefficients on the next level; (b) a sample face image with its wavelet coefficients at different decomposing levels.} 
\label{fig:waveletPacketTransform}
\end{figure}

\subsection{Wavelet-based Discriminator}
To force the generator to absorb the semantic information of the input face image, a conditional discriminator is employed. The discriminator has two main functions: 1) distinguish synthesized face images from generic ones; 2) check whether the attribute of each generation result is faithful to that of the corresponding input.

To be specific, considering the fact that typical signs of aging, such as wrinkles, laugh lines, and eye bags, could be regarded as local image textures, we adopt wavelet packet transform (WPT, see Fig.~\ref{fig:waveletPacketTransform}) to capture age-related textural features.
Specifically, multi-level WPT is performed to provide a more comprehensive analysis of textures in the given image, and wavelet coefficients at each decomposing level are fed into a convolutional pathway of the discriminator. 
Note that this is different from~\cite{li2018global}, since wavelet coefficients are only used for discrimination in our work and no prediction or reconstruction is involved.

To make the discriminator gain the ability of telling whether attributes are preserved in generated images, the input attribute vector is also replicated and concatenated to the output of an intermediate convolutional block of each pathway.
At the end of the discriminator, same-sized outputs of all pathways are fused to form a single tensor, and adversarial loss is then estimated against the label tensor.

Compared to extracting multi-scale features by a sequence of convolutional layers as in~\cite{yang2017learning}, the advantage of using WPT is that the computational cost is significantly reduced since calculating wavelet coefficients could be regarded as forwarding through a single convolutional layer. 
Therefore, WPT greatly reduces the number of convolutions performed in each forwarding process.
Although this part of the model has been simplified, it still takes the advantage of multi-scale image texture analysis, which is helpful in improving the visual fidelity of generated images. 

\subsection{Overall Objective Functions}

Training of GAN model simulates the process of optimizing a minimax-max two-player game between the generator $G$ and the discriminator $D$. 
Unlike regular GANs~\cite{goodfellow2014generative}, we adopt least square loss instead of negative log likelihood loss for that margins between generated samples and the decision boundary in the feature space are also minimized, which further improves the quality of synthesized images~\cite{mao2017least}. 
Practically, we pair up young face images $x_i$ and their corresponding attribute vectors $\alpha_i$ of dimension $p$, denoted as $(x_i, \alpha_i)\sim P_{young}(x, \alpha)$, and take them as input to the model. 
Only generic aged faces with attributes same as the input,~\ie $(x_i, \alpha_i)\sim P_{old}(x, \alpha_i)$, are considered as positive samples, and real young faces,~\ie $(x_i, \alpha_i)\sim P_{young}(x, \alpha)$, are regarded as negative samples to help $D$ gain discriminating ability on aging effects.

Mathematically, the objective function for $G$ and $D$ could be written as follows,
\begin{equation}
L_{GAN}(G)=\mathbb{E}_{(x_i, \alpha_i)\sim P_{young}(x, \alpha)}[(D(G(x_i, \alpha_i), \alpha_i)-1)^2]
\end{equation}
\begin{equation}
\begin{split}
L_{GAN}(D)=&\,\mathbb{E}_{(x_i, \alpha_i)\sim P_{old}(x, \alpha_i)}[(D(x_i, \alpha_i)-1)^2]+\\
           &\,\mathbb{E}_{(x_i, \alpha_i)\sim P_{young}(x, \alpha)}D(G(x_i, \alpha_i), \alpha_i)^2+\\
           &\,\mathbb{E}_{(x_i, \alpha_i)\sim P_{young}(x, \alpha)}D(x_i, \alpha_i)^2
\end{split}
\end{equation}
where $P_{young}$ and $P_{old}$ denote the distribution of generic face images of young and old subjects, respectively. 

In addition, pixel loss and identity loss are adopted to maintain consistency in both image-level and personalized feature-level. 
To be specific, we utilize the VGG-Face descriptor~\cite{parkhi2015deep}, denoted by $\phi$, to extract the identity related semantic representation of a face image. These two loss terms could be formulated as,
\begin{equation}
L_{pix}=\mathbb{E}_{(x_i, \alpha_i)\sim P_{young}(x, \alpha)}||G(x_i, \alpha_i)-x_i||_F^2
\end{equation}
\begin{equation}
L_{id}=\mathbb{E}_{(x_i, \alpha_i)\sim P_{young}(x, \alpha)}||\phi(G(x_i, \alpha_i))-\phi(x_i)||_F^2
\end{equation}

In conclusion, overall objective functions of the proposed model could be written as follows,
\begin{equation}
L_G=L_{GAN}(G)+\lambda_{pix}L_{pix}+\lambda_{id}L_{id} \\
\end{equation}
\begin{equation}
L_D=L_{GAN}(D)
\end{equation}
where $\lambda_{id}$ and $\lambda_{pix}$ are coefficients balancing the importance of critics on identity and pixels, respectively. We optimize the model by minimizing $L_G$ and $L_D$ alternatively until the optimality is reached.

\begin{figure*}[ht]
\centering\includegraphics[width=0.9\linewidth]{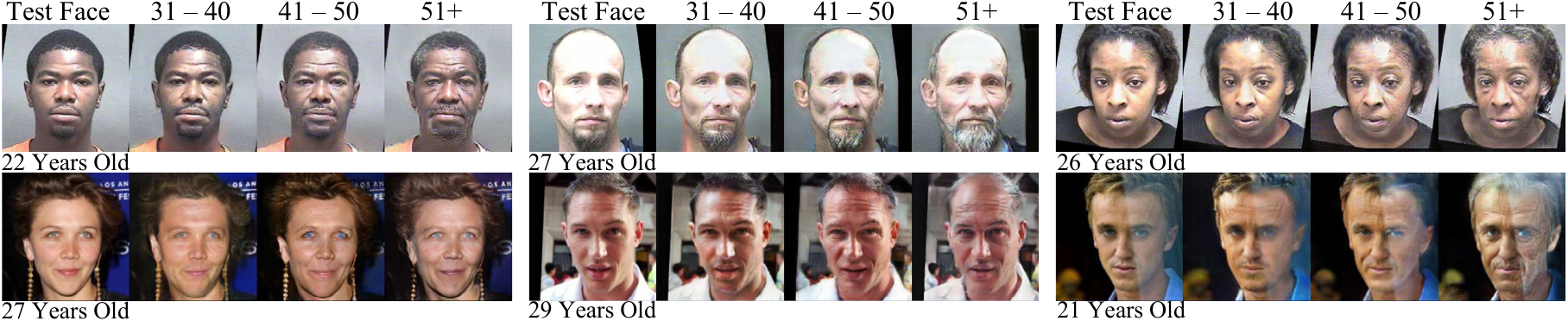}
\caption{Sample results on Morph (first row) and CACD (second row). The first image in each result is the input test face image and subsequent 3 images are synthesized elderly face images of the same subject in age group 31-40, 41-50 and 51+, respectively.}
\label{fig:sampleResults}
\end{figure*}

\begin{figure*}[ht]
\centering\includegraphics[width=0.80\linewidth]{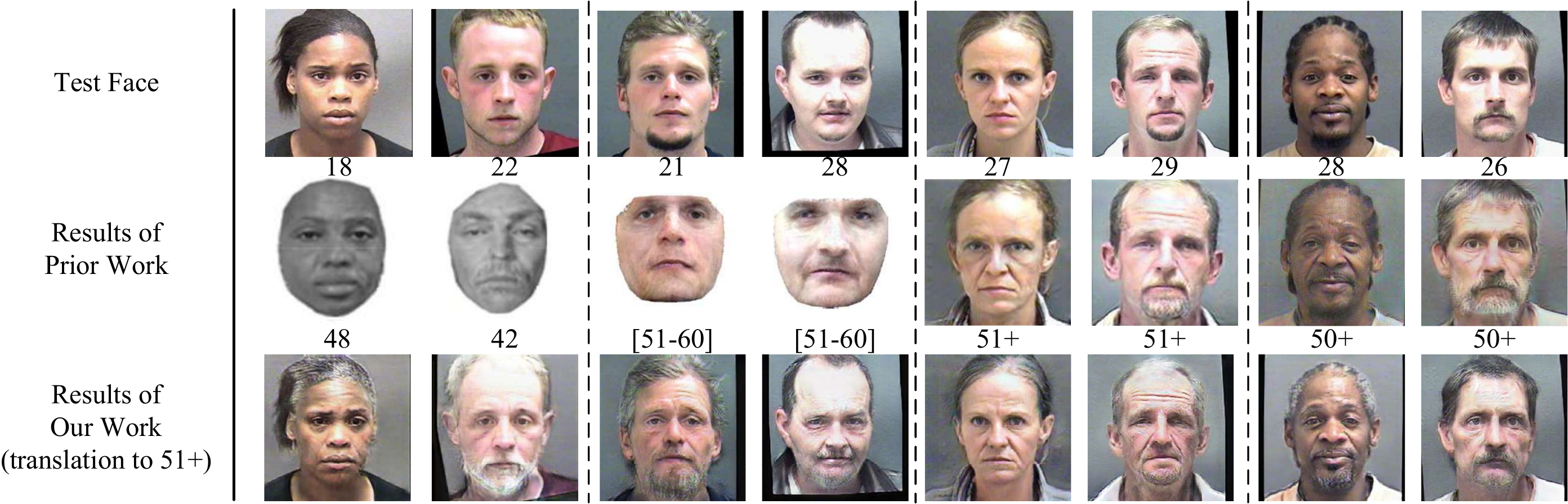}
\caption{Performance comparison with prior work on Morph (zoom in for a better view of the aging details). The second row shows the results of prior work, where four methods are considered and two sample results are presented for each. These four methods are (from left to right): CONGRE~\cite{suo2012concatenational}, HFA~\cite{yang2016face}, GLCA-GAN~\cite{li2018global}, and PAG-GAN~\cite{yang2017learning}. The last row shows the results of our method.}
\label{fig:compareResults}
\end{figure*}

\section{Experiments}

\subsection{Dataset}
\textbf{MORPH}~\cite{ricanek2006morph} is a large aging dataset containing 55,000 face images of more than 13,000 subjects. 
Data samples in MORPH are color images of near-frontal faces exhibiting neutral expressions under uniform and moderate illumination with simple background. 
\textbf{CACD}~\cite{chen2015face} contains 163,446 face images of 2,000 celebrities captured in much less controlled conditions. Besides large variations in pose, illumination, and expression (PIE variations), images in CACD are collected via Google Image Search, making it a very challenging dataset due to the mismatching between actual face presented in each image and associated labels provided (name and age).

As for facial attributes, MORPH provides researchers with labels including age, gender, and race for each image. We choose `gender' and `race' to be the attributes that are required to be preserved, since these two attributes are guaranteed to remain unchanged during natural aging process, and are relatively objective compared to attributes such as `attractive' or `chubby' used in popular facial attribute dataset CelebA~\cite{liu2015faceattributes}.
For CACD, since face images with race other than `white' only takes a small portion of the entire dataset, we only select `gender' as the attribute to preserve. 
To be specific, we go through the name list of the celebrities and label the corresponding images accordingly. This introduces noise in gender labels due to the mismatching between the annotated name and the actual face presented in each image, which further increases the difficulty for our method to achieve good performance on this dataset.
It is worthwhile to note that the proposed model is highly expandable, as researchers may choose whatever attributes to preserve simply by incorporating them in the conditional facial attribute vector and arrange training images pairs accordingly.

\subsection{Implementation Details}
All face images are cropped and aligned according to the five facial landmarks detected by MTCNN~\cite{zhang2016joint}. 
Following the convention in~\cite{yang2017learning,li2018global}, we divide the face images into four age groups,~\ie, 30-, 31-40, 41-50, 51+, and only consider translations from 30- to the other three age groups.
To evaluate the performance of the proposed method objectively, all metric measurements are conducted via stable public APIs of Face++~\cite{face2018toolkit}. 
Thresholds adopted in our face verification experiments (threshold=76.5, FAR=1e-5) are the same as those used in~\cite{yang2017learning}. Therefore, quantitative results of our experiments are comparable to those reported in~\cite{yang2017learning}.

We choose Adam to be the optimizer of both $G$ and $D$ with learning rate and batch-size set to $1e^{-4}$ and 16, respectively. 
Pixel-level critic is applied every 5 iterations, and $D$ is updated at every iteration. 
As for trade-off parameters, $\lambda_{pix}$ and $\lambda_{id}$ are firstly set to make $L_{pix}$ and $L_{id}$ to be of the same order of magnitude as $L_{GAN}(G)$, and then divided by 10 to emphasize the importance of the adversarial loss.
All experiments are conducted under 5-fold cross validation on a Nvidia Titan Xp GPU.

\begin{table*}[ht]
\centering
\caption{Age estimation results on Morph and CACD (differences of mean ages are measured in absolute value).}
\begin{tabular} {@{}l RRR c l RRR@{}}
\toprule
          \multicolumn{4}{c}{Morph}                       &\phantom{ab} & \multicolumn{4}{c}{CACD}                 \\
          \cmidrule{1-4}                                                  \cmidrule{6-9}
Age group & 31 - 40 & 41 - 50 & 51 +                      &\phantom{ab} & Age group & 31 - 40 & 41 - 50 & 51 +  \\
\midrule
          \multicolumn{4}{c}{Estimated Age Distributions} &\phantom{ab} & \multicolumn{4}{c}{Estimated Age Distributions} \\
          \cmidrule{1-4}                                                  \cmidrule{6-9}                                                  
Generic   & 38.60         & 47.74       & 57.25           &\phantom{ab} & Generic   & 38.51         & 46.54         & 53.39 \\
Synthetic & 38.47         & 47.55       & 56.57           &\phantom{ab} & Synthetic & 38.88         & 47.42         & 54.05 \\
          \multicolumn{4}{c}{Difference of mean ages}     &\phantom{ab} & \multicolumn{4}{c}{Difference of mean ages} \\
          \cmidrule{1-4}                                                  \cmidrule{6-9}                                                  
CAAE      & 10.08         & 15.49         & 21.42         &\phantom{ab} & CAAE      & 5.76          & 11.53         & 17.93          \\
GLCA-GAN  & 0.23          & 3.61          & 8.61          &\phantom{ab} & GLCA-GAN  & 1.72          & 2.07          & 2.85           \\
PAG-GAN   & 0.38          & 0.52          & 1.48          &\phantom{ab} & PAG-GAN   & 0.70          & \textbf{0.22} & \textbf{0.57}  \\ 
Ours      & \textbf{0.13} & \textbf{0.19} & \textbf{0.68} &\phantom{ab} & Ours      & \textbf{0.37} & 0.58          & 0.66           \\
\bottomrule
\end{tabular}
\label{table:AgeAcc}
\end{table*}

\begin{figure}[ht]
\centering\includegraphics[width=1.0\linewidth]{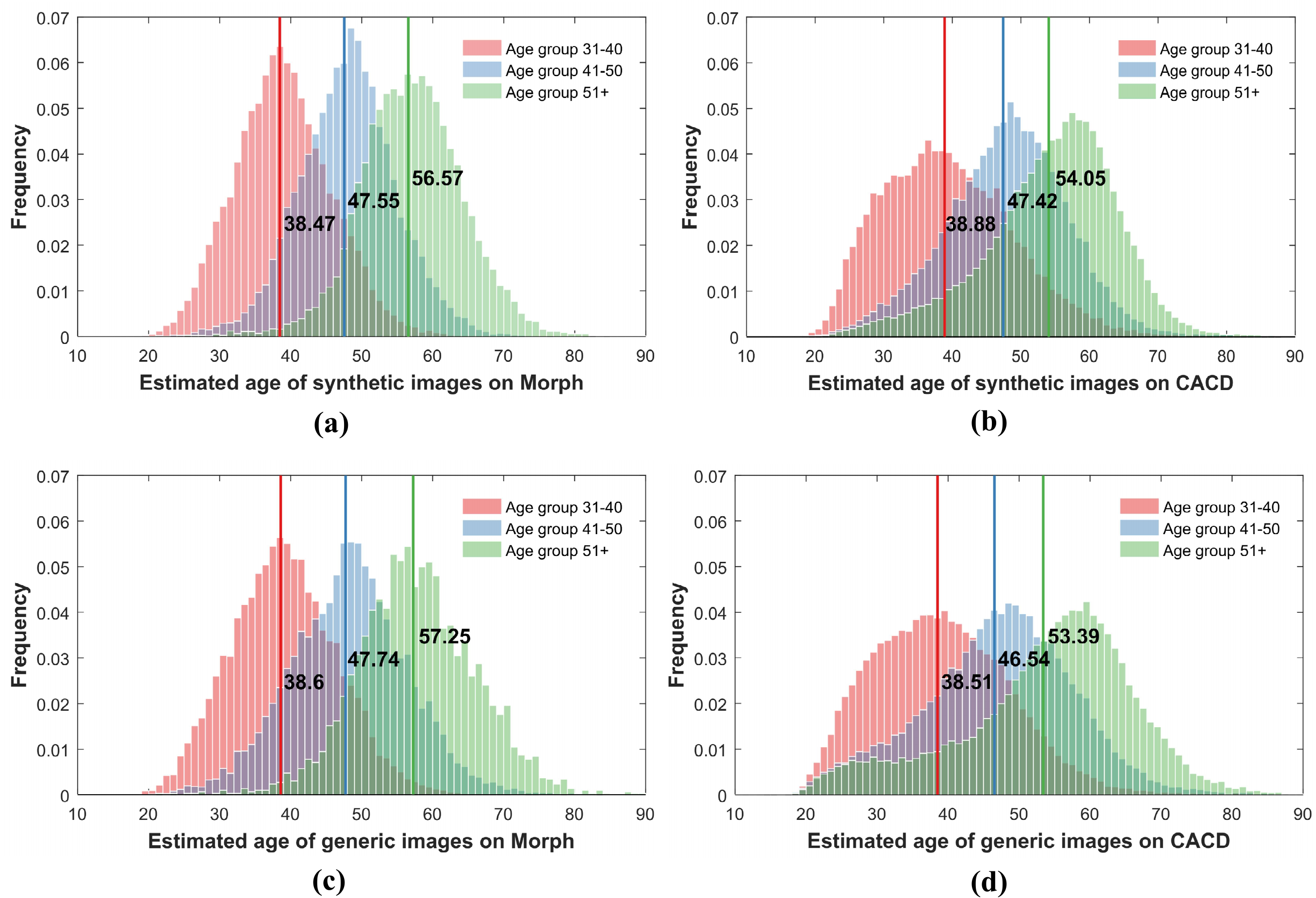}
\caption{Distributions of the estimated ages. (a) synthetic faces on Morph; (b) synthetic faces on CACD; (c) generic faces on Morph; (d) generic faces on CACD.}
\label{fig:histAgeEst}
\end{figure}

\subsection{Qualitative Results of Face Aging}
Sample results on Morph and CACD are shown in Fig.~\ref{fig:sampleResults}. It is clear that our method is able to simulate translations between age groups and synthesize elderly face images with high visual fidelity. In addition, our method is robust to variations in terms of race, gender, expression, and occlusion.

Performance comparison with prior work on Morph is shown in Fig.~\ref{fig:compareResults}. Traditional face aging methods, CONGRE~\cite{suo2012concatenational} and HFA~\cite{yang2016face}, only render subtle aging effects within tight facial area, which fails to accurately simulate the aging process.
In contrast, GAN-based methods, GLCA-GAN~\cite{li2018global} and GAN with pyramid architecture proposed in~\cite{yang2017learning}, referred to as PAG-GAN, have achieved significant improvement on the quality of generation results. 
However, our method further generates face images of higher resolution ($2\times$) with enhanced details compared to GLCA-GAN, and reduces ghosting artifacts in the results compared to PAG-GAN (\eg finer details of hair and beard).

\subsection{Aging Accuracy and Identity Preservation}
In this subsection, we report evaluation results on aging accuracy and identity preservation. The performance of the proposed model is compared with previously state-of-the-art methods CAAE~\cite{zhang2017age}, GLCA-GAN~\cite{li2018global} and PAG-GAN~\cite{yang2017learning} to demonstrate the effectiveness. 

\textbf{Aging Accuracy:} Age distributions of both generic and synthetic faces in each age group are estimated, where less discrepancy between real and fake images indicates more accurate simulation of aging effects. 
On Morph and CACD, face images of age under or equal to 30 are considered as testing samples, and their corresponding aged faces in the other three age groups are synthesized. 
We estimated the apparent age of both generation results and natural face images in the dataset using Face ++ APIs for fair comparison.

Age estimation results on Morph and CACD are shown in Table~\ref{table:AgeAcc} and Fig.~\ref{fig:histAgeEst}. 
We compare our method with previous works in terms of differences between mean ages. 
On Morph, it could be seen that estimated age distributions of synthetic elderly face images well match that of natural images for all age groups. Our method consistently outperforms other approached in all three aging processes, demonstrating the effectiveness of our method. 
Signs of aging in results of CAAE are not obvious enough, leading to large age estimation errors. 
On CACD, due to the existence of mismatching between face images and associated labels, slight performance drop could be observed. Still, the proposed method achieves results comparable to previous state-of-the-art. 
This shows that our method is relatively robust to noise in attribute labels and thus lower the requirement on the accuracy of the prior attribute detection process.


\begin{table*}[ht]
\centering
\caption{Face verification results on Morph and CACD.}
\begin{tabular} {@{}l RRR c l RRR@{}}
\toprule
             \multicolumn{4}{c}{Morph}                        &\phantom{ab}  &          \multicolumn{4}{c}{CACD}                 \\
             \cmidrule{1-4}                                                             \cmidrule{6-9}                              
Age group  & 31 - 40        & 41 - 50        & 51 +           &\phantom{ab}  & Age group& 31 - 40        & 41 - 50        & 51 +   \\
\midrule
           \multicolumn{4}{c}{Verification Confidence}        &\phantom{ab}  &          \multicolumn{4}{c}{Verification Confidence} \\
             \cmidrule{1-4}                                                             \cmidrule{6-9}
30 -       & 95.77          & 94.64          & 87.53          &\phantom{ab}  & 30 -     & 93.67          & 91.54          & 90.32  \\
31 - 40    & -              & 95.47          & 89.53          &\phantom{ab}  & 31 - 40  & -              & 91.74          & 90.54  \\
41 - 50    & -              & -              & 90.50          &\phantom{ab}  & 41 - 50  & -              & -              & 91.12  \\
           \multicolumn{4}{c}{Verification Rate (\%)}         &\phantom{ab}  &          \multicolumn{4}{c}{Verification Rate (\%)}          \\
           \cmidrule{1-4}                                                               \cmidrule{6-9}
CAAE     & 15.07           & 12.02           & 8.22           &\phantom{ab}  & CAAE     & 4.66            & 3.41            & 2.40           \\
GLCA-GAN & 97.66           & 96.67           & 91.85          &\phantom{ab}  & GLCA-GAN & 97.72           & 94.18           & 92.29          \\
PAG-GAN  & \textbf{100.00} & 98.91           & 93.09          &\phantom{ab}  & PAG-GAN  & \textbf{99.99}  & \textbf{99.81}  & 98.28          \\
Ours     & \textbf{100.00} & \textbf{100.00} & \textbf{98.26} &\phantom{ab}  & Ours     & 99.76           & 98.74           & \textbf{98.44} \\
\bottomrule
\end{tabular}
\label{table:IdPreserve}
\end{table*}


\begin{table*}[ht]
\centering
\caption{Facial attributes preservation rates for `Gender' and `Race' on Morph and CACD.}
\begin{tabular} {@{}l RRR c RRR c RRR@{}}
\toprule
           & \multicolumn{7}{c}{Preservation Rate (\%) of `Gender'}                                                           &\phantom{a} & \multicolumn{3}{c}{Preservation Rate (\%) of `Race'} \\
             \cmidrule{2-8}                                                                                                                  \cmidrule{10-12}
           & \multicolumn{3}{c}{Morph}                        &\phantom{a} & \multicolumn{3}{c}{CACD}                         &\phantom{a} & \multicolumn{3}{c}{Morph} \\
             \cmidrule{2-4}                                                  \cmidrule{6-8}                                                  \cmidrule{10-12}
Age group  & 31 - 40        & 41 - 50        & 51 +           &\phantom{a} & 31 - 40        & 41 - 50        & 51 +           &\phantom{a} & 31 - 40        & 41 - 50        & 51 +           \\
\midrule
GLCA-GAN   & 96.30          & 95.43          & 95.77          &\phantom{a} & 87.27          & 86.79          & 85.89          &\phantom{a} & 91.79          & 89.52          & 89.34          \\
PAG-GAN    & 95.96          & 93.77          & 92.47          &\phantom{a} & 83.97          & 81.28          & 70.05          &\phantom{a} & 95.83          & 88.51          & 87.98          \\
Ours       & \textbf{97.37} & \textbf{97.21} & \textbf{96.07} &\phantom{a} & \textbf{90.71} & \textbf{87.63} & \textbf{87.19} &\phantom{a} & \textbf{95.86} & \textbf{94.10} & \textbf{93.22} \\
\bottomrule
\end{tabular}
\label{table:AttPreserve}
\end{table*}

\textbf{Identity Preservation:} Face verification experiments are conducted to check whether the identity information has been preserved during the face aging process. 
Similar to previous literature, comparisons between synthetic elderly face images from different age groups of the same subject are also conducted to inspect if the identity information is consistent among three separately trained age mappings.

Results of face verification experiments are shown in Table~\ref{table:IdPreserve}. 
On Morph, our method achieves the highest verification rate on all three translations and outperforms other approaches by a clear margin, especially in the hardest case (from 30- to 51+). This demonstrates that the proposed method successfully achieves identity permanence during face aging. On the more challenging dataset CACD containing mismatched labels, the performance of our method is comparable to PAG-GAN with minor difference.
Notably, as the time interval between two face images of a single subject increases, both verification confidence and accuracy decrease, which is reasonable as greater changes in facial appearance may occur as more time elapsed. 

\subsection{Facial Attribute Consistency}
We evaluate the performance of facial attribute preservation by comparing facial attributes estimated before and after age progression, and results are listed in Table~\ref{table:AttPreserve}. 
On Morph, facial attributes of the majority of testing samples (up to 97.37\% for `gender' and 95.86\% for `race') are well preserved in the aging process. In addition, our method outperforms both GLCA-GAN and PAG-GAN by clear margins on translations to all age groups. 
On CACD, due to the influence of mistakenly labeled data samples, clear performance drop could be observed compared to the results on Morph. However, our method still gives better performance on facial attributes preservation than other methods. 
The advantage of our method in preserving the `gender' attribute becomes greater as the age gap increases, and finally reaches 17.14\% (87.19\% over 70.05\%) when translating to the oldest age group 51+. 
From Table~\ref{table:AttPreserve}, we could conclude that undesired changes of facial attributes are more likely to happen as the age gap increases, and incorporating conditional information is beneficial for maintaining consistency of target facial attributes in the aging process.

\begin{figure}[ht]
\centering\includegraphics[width=0.9\linewidth]{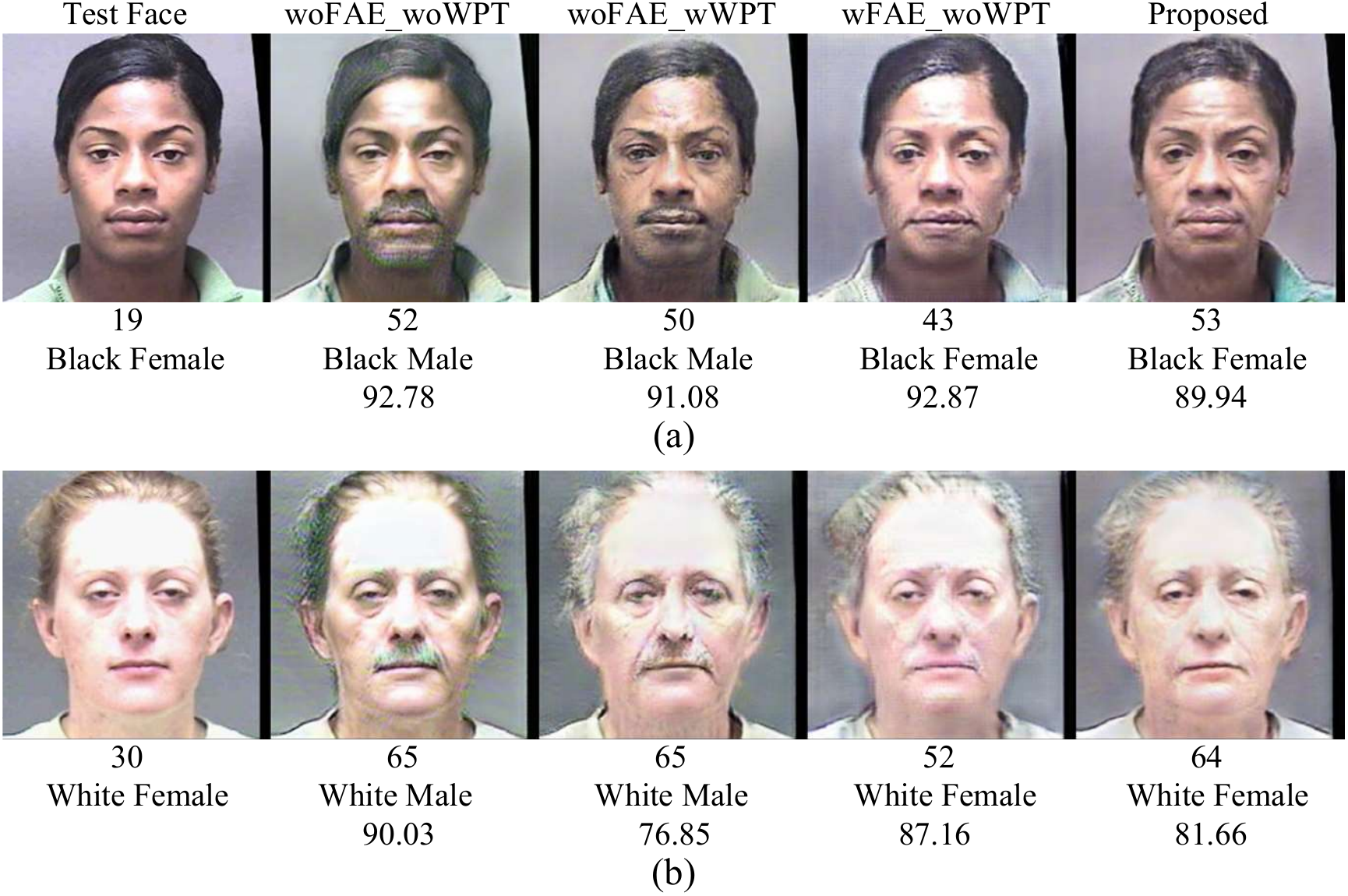}
\caption{Sample visual results of the ablation study. 
For each face, the estimated age (first row) and detected attributes (second row) are listed underneath. Values in the last row are face verification confidence between generation results and the test face.}
\label{fig:ablationStudyVis}
\end{figure}

\subsection{Ablation Study}

In this part, experiments are conducted to fully explore the contribution of facial attribute embedding (FAE) and wavelet packet transform (WPT) in simulating accurate age translations. 
We investigate the impact of including/excluding attribute embedding (w/wo FAE) and wavelet packet transform (w/wo WPT) on age distribution, face verification rate, and attribute preservation rate. 
All experiments in this subsection are conducted only on Morph as labels are noisy on CACD dataset.

\begin{table*}[ht]
\centering
\caption{Comparison of results on facial attribute preservation and aging accuracy between variants of the proposed model (differences of mean ages are measured in absolute value).}
\begin{tabular}{@{}l RRR c RRR c RRR@{}}
\toprule
              & \multicolumn{3}{c}{Gender Preservation Rate (\%)} &\phantom{a} & \multicolumn{3}{c}{Race Preservation Rate (\%)} &\phantom{a} & \multicolumn{3}{c}{Deviation of Estimated Ages}\\
                \cmidrule{2-4}                                                   \cmidrule{6-8}                                                 \cmidrule{10-12}
Age group     & 31-40 & 41-50 & 51+                               &            & 31-40 & 41-50 & 51+                             &            & 31-40 & 41-50 & 51+ \\
\midrule                
woFAE / woWPT & 95.72 & 94.21 & 93.60                             &            & 95.04 & 93.55 & 90.83                           &            & 0.44  & 1.72  & 3.03 \\
woFAE / wWPT  & 96.15 & 94.90 & 93.61                             &            & 93.89 & 88.63 & 90.21                           &            & 0.68  & 0.41  & 2.31 \\
wFAE / woWPT  & 97.21 & 96.91 & 95.85                             &            & 95.22 & \textbf{94.35} & 91.43                  &            & 0.82  & 0.52  & 4.82 \\
Ours          & \textbf{97.37} & \textbf{97.21} & \textbf{96.07} & & \textbf{95.86} & 94.10 & \textbf{93.22}     &            & \textbf{0.13} & \textbf{0.19} & \textbf{0.68} \\
\bottomrule
\end{tabular}
\label{table:AgeAccAttPreAblation}
\end{table*}


\begin{table}[ht]\centering
\caption{Face verification rates (\%) of variants of the proposed model on Morph}
\begin{tabular}{@{}l rrr@{}}
\toprule
Age group & 31-40 & 41-50 & 51+ \\
\midrule
woFAE / woWPT & 100.00  & 100.00  & 99.92 \\
woFAE / wWPT  & 100.00  & 99.88   & 98.06 \\
wFAE / woWPT  & 100.00  & 100.00  & 98.86 \\
Ours          & 100.00  & 100.00  & 98.26 \\
\bottomrule
\end{tabular}
\label{table:IdPreAblation}
\end{table}

Visual illustrations of face images generated by variants of the proposed model are shown in Fig.~\ref{fig:ablationStudyVis}. It is clear that when both FAE and WPT are not involved (woFAE{\_}woWPT), generation results suffer from severe ghosting artifacts. 
Due to the intrinsic matching ambiguity of unpaired training data, the model without FAE mistakenly attaches moustache to the input female face image to show the aging effect. 
Notably, growing a moustache does not decrease the face verification confidence, as the generated face image still shares similar identity-related features with the input. 
This again confirms our observation that enforcing identity consistency is insufficient to obtain satisfactory face aging results.

On the contrary, incorporating FAE suppresses the undesired facial attribute drift by reducing the matching ambiguity.
To be specific, in Fig.~\ref{fig:ablationStudyVis}, there is no more moustache in generation results after adopting FAE thus facial attribute consistency is achieved. 
Unfortunately, removing moustache also wipes out aging-related textural details (wrinkles, laugh lines, and eye bags), leading to relatively inaccurate aging results (much younger than expected).

To solve this issue and generate more visually plausible face images with vivid signs of aging, WPT is employed as the initial layer of the discriminator. 
The contribution of WPT could be easily seen by comparing the results obtained under setting `woFAE / woWPT' and `woFAE / wWPT', as well as `wFAE / woWPT' and `Ours'. 
Although results obtained under setting `woFAE / wWPT' still suffer from wrong facial attributes, ghosting artifacts are significantly alleviated and lifelike aging effects are clearly observed. 

Quantitative results for ablation study are shown in Table~\ref{table:AgeAccAttPreAblation} and~\ref{table:IdPreAblation}.
According to results in Table~\ref{table:AgeAccAttPreAblation}, introducing facial attribute embedding (wFAE) increases preservation rates for both `gender' and `race' under all three age mappings, especially in the case of translating to 51+. 
This proves the effectiveness of attribute embedding as it aligns unpaired age data in terms of facial attributes and thus reduces the intrinsic ambiguity in data mapping. 

In addition, it is clear that adopting WPT reduces the discrepancies between age distributions of generic and synthetic images in all cases. 
However, WPT provides little help in maintaining facial attribute consistency.
This is because WPT only captures feature based on low-level visual data and could not bridge the semantic gap, so that the framework still suffers from mismatched data samples.

Combining results in Table~\ref{table:AgeAccAttPreAblation} and~\ref{table:IdPreAblation}, it could be seen that while attribute preservation rates still have room for improvement, verification rates are about to reach perfection.
This observation validates our statement that identity preservation does not guarantee that facial attributes remain stable during the aging process.
Therefore, besides constraints on identity, supervision on facial attributes are also helpful to reduce the intrinsic matching ambiguity of unpaired data and achieve satisfactory face aging results.

\section{Conclusion}
In this paper, we propose a GAN-based framework to synthesize aged face images. Due to the ineffectiveness of identity constraints in reducing the matching ambiguity of unpaired aging data, we propose to employ facial attributes to tackle this issue. 
Specifically, we embed facial attribute vectors to both the generator and discriminator to encourage generated images to be faithful to facial attributes of the corresponding input image. 
To further improve the visual fidelity of generated face images, wavelet packet transform is introduced to extract textual features at multiple scales efficiently. 
Extensive experiments are conducted on Morph and CACD, and qualitative results demonstrate that our method could synthesize lifelike face images robust to both PIE variations and noisy labels. 
Furthermore, quantitative results obtained via public APIs validate the effectiveness of the proposed method in aging accuracy as well as identity and attribute preservation.

\textbf{Acknowledgements.} This work is supported by the National Natural Science Foundation of China (Grant No. 61702513, U1836217, 61427811).

{\small
\bibliographystyle{ieee}
\bibliography{egbib}
}

\end{document}